\pdfoutput=1

\documentclass[11pt]{article}

\usepackage[]{ACL2023}
\usepackage{soul}
\usepackage{times}
\usepackage{latexsym}

\usepackage[T1]{fontenc}

\usepackage[utf8]{inputenc}

\usepackage{microtype}

\usepackage{inconsolata}

\usepackage{booktabs}

\usepackage{multirow}

\usepackage{graphicx}

\usepackage{xcolor}

\usepackage{caption}
\usepackage{subcaption}
\usepackage{ upgreek }

\newcommand{\zscda}{CDA-$\{\mathrm{en}\}$}
\newcommand{\monocda}{CDA-$\{l\}$}
\newcommand{\fscda}{CDA-$\{l,\mathrm{en}\}$} 
\newcommand{\indcda}{CDA-$\mathcal{L}  \setminus \{\mathrm{en}\}$}

\title{On Evaluating and Mitigating Gender Biases in Multilingual Settings}

\author{Aniket Vashishtha\thanks{\hspace{0.1cm} Equal contribution} \quad
  Kabir Ahuja\footnotemark[1] \quad
  Sunayana Sitaram \\
 Microsoft Research India \\ {\tt \small \{t-aniketva,t-kabirahuja,sunayana.sitaram\}@microsoft.com}
}

\begin{document}
\maketitle
\begin{abstract}
While understanding and removing gender biases in language models has been a long-standing problem in Natural Language Processing, prior research work has primarily been limited to English. In this work, we investigate some of the challenges with evaluating and mitigating biases in multilingual settings which stem from a lack of existing benchmarks and resources for bias evaluation beyond English especially for non-western context. In this paper, we first create a benchmark for evaluating gender biases in pre-trained masked language models by extending DisCo to different Indian languages using human annotations. We extend various debiasing methods to work beyond English and evaluate their effectiveness for SOTA massively multilingual models on our proposed metric. Overall, our work highlights the challenges that arise while studying social biases in multilingual settings and provides resources as well as mitigation techniques to take a step toward scaling to more languages.
\end{abstract}

\section{Introduction}
Large Language Models (LLMs) \cite{devlin-etal-2019-bert, brown-etal-2020-language, raffel-etal-2020-exploring} have obtained impressive performance on a wide range of NLP tasks showing great potential in several downstream applications for real world impact. However, these models have shown to be prone to picking up unwanted correlations and stereotypes from the pre-training data \cite{sheng-etal-2019-woman, kurita-etal-2019-measuring, hutchinson-etal-2020-social} which, can perpetuate harmful biases for people belonging to marginalized groups. While there has been a great deal of interest in understanding and mitigating such biases in LLMs \cite{nadeem-etal-2021-stereoset, schick-etal-2021-self, meade-etal-2022-empirical}, the focus of such studies has primarily been on English.

While Massively Multilingual Language Models \cite{devlin-etal-2019-bert, conneau-etal-2020-unsupervised, xue-etal-2021-mt5}, have shown impressive performances across a wide range of languages, especially with their surprising effectiveness at zero-shot cross-lingual transfer, there still exists a lack of focused research to evaluate and mitigate the biases that exist in these models. This  can lead to a lack of inclusive and responsible technologies for groups whose native language is not English and can also lead to the dissemination of stereotypes and the widening of existing cultural gaps.

Past work on evaluating and mitigating biases in multilingual models has mostly been concerned with gender bias in cross-lingual word embeddings \cite{zhao-etal-2020-gender, bansal-etal-2021-debiasing} which fails to account for contextual information \cite{kurita-etal-2019-measuring, delobelle-etal-2022-measuring}, making them unreliable for LLMs. Other methods for estimating biases in contextualized representations involve Multilingual Bias Evaluation \cite[MBE]{kaneko-etal-2022-gender}, which utilizes parallel translation corpora in different languages that might lack non-western cultural contexts \cite{talat-etal-2022-reap}. For debiasing LLMs, \citet{lauscher-etal-2021-sustainable-modular} proposed an adapter \cite{houlsby-etal-2019-parameter} based approach. However, the biases are measured in the word representations and only English data was used for debiasing, missing out on cultural context for other languages.

To address these concerns, we make the following key contributions in our work. \textit{First}, we extend the DisCo metric \cite{webster2020measuring} by creating human-corrected templates for 6 Indian languages. 
DisCo takes sentence-level context while measuring bias and our templates are largely culturally agnostic making them more generally applicable. \textit{Second}, we extend existing debiasing strategies like Counterfactual Data Augmentation \cite{zhao-etal-2018-gender} and Self-Debiasing \cite{schick-etal-2021-self} to mitigate gender biases across languages in Masked Language Models (MLMs). 



\textit{Finally}, we also evaluate the transferability of debiasing MLMs from one source language to other target languages and observe limited transfer from English to languages lacking western context. However, we do observe that typologically and culturally similar languages aid each other in reducing gender bias. While there have been multiple studies on measuring biases in multilingual models, previous work has not explored mitigating gender biases from these models on multiple languages and studying the transferability of debiasing across different languages. This is especially true while using non-embedding based approaches for evaluation and debiasing. To the best of our knowledge, ours is the first work to debias multilingual LLMs for different languages and measure the cross-lingual transfer for gender bias mitigation. To encourage future research in this area, we will release our code and datasets publically\footnote{\url{https://aka.ms/multilingual-bias}}.

\section{Measuring Bias in Multilingual Models}

In this section, we describe the benchmarks to evaluate biases in MLMs across different languages. Since most existing benchmarks for bias evaluation in contextualized representations are designed for English, we discuss our multilingual variant of DisCo and the recently proposed MBE metric.


\subsection{Multilingual DisCo}


Discovery of Correlations (DisCo) is a template-based metric that measures unfair or biased associations of predictions of an MLM to a particular gender. It follows a slot-filling procedure where for each template, predictions are made for a masked token, which are evaluated to assess whether there is a statistically significant difference in the top predictions across male and female genders. For calculating the bias score using DisCo, a $\upchi^2$ test is performed to reject the null hypothesis (with a p-value of 0.05) that the model has the same prediction rate with both male and female context. We use the modified version of the metric from \cite{delobelle-etal-2022-measuring} that measures the fraction of slot-fills containing predictions with gendered associations (fully biased model gets a score of 1, and fully unbiased gets a score of 0).



We extend the \textbf{Names} variant of DisCo, as personal names can act as representatives for various socio-demographic attributes to capture cultural context \cite{sambasivan-etal-2021-imagining}. Especially for India, surnames are a strong cultural identifier. Majority Indian surnames are typically an identifier of belonging to a particular caste, religion and culture. We use surnames from specific cultures which speak the languages for which we prepare the name pairs for. We further use these surnames to filter out personal first names for both male and female from an open-source Indian names list containing a large number of popular Indian names (details in Appendix \ref{app:data_const}) and word-translated the names from English to the corresponding languages, to be used for slot-filling.
Further, unlike nouns and pronouns which might be gender-neutral in some languages, names are indicative of gender to a large extent across cultures.



\begin{figure}
    \centering
    \includegraphics[width=.49\textwidth]{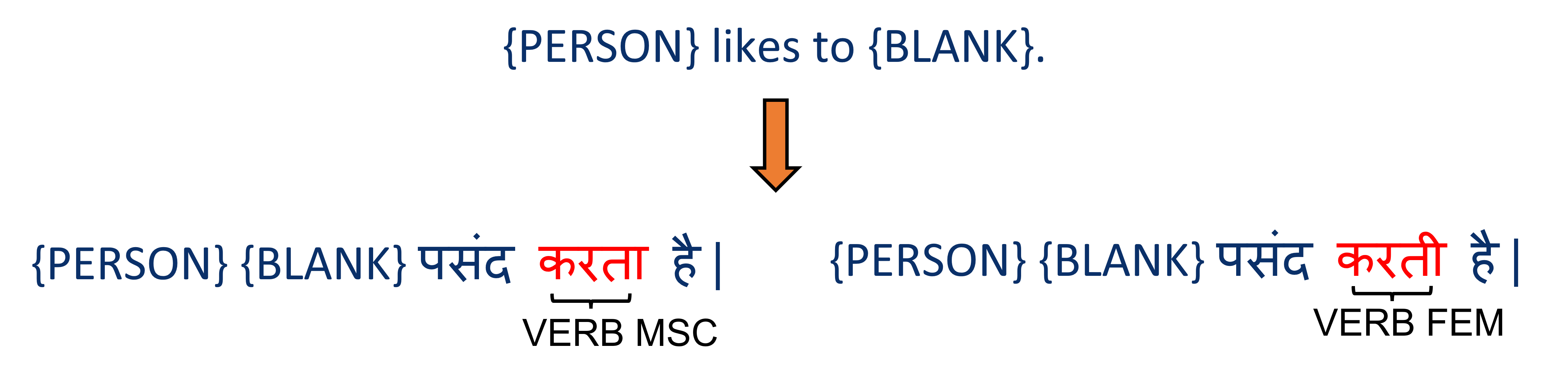}
    \caption{Example template translation for ``\textit{\{PERSON\} likes to \{BLANK\}}'' in Hindi for creation of our multilingual dataset.}
    \label{fig:template_translate}
\end{figure}

\noindent
\textbf{Dataset Construction}: We start with the 14 templates provided in \citet{webster2020measuring} and translate them using Bing translation API \footnote{\url{https://www.microsoft.com/en-us/translator/}} to 6 Indian languages of varying resources. We use the Class taxonomy from \cite{joshi-etal-2020-state} to characterize language resources, where Class 5 represent high resource and Class-0 for lowest resource languages. Our set of Indian Languages contain Class 4 language Hindi (\textit{hi}); Class 3 language Bengali (\textit{bn}); Class 2 languages Marathi (\textit{mr}) and Punjabi (\textit{pa}); and Class 1 language Gujarati (\textit{gu}). A challenge while transferring templates from English to these languages is that, unlike English, a common template might not be applicable to both genders. For eg. the template ```\textit{\{PERSON\} likes to \{BLANK\}}''', will have different translations in Hindi, depending upon the gender of the slot fill for \{PERSON\}, as Hindi has gendered verbs. Hence, during translation we first filled the \textit{\{PERSON\}} slot with a male and a female name to obtain two templates corresponding to each gender (see Figure \ref{fig:template_translate}). All the translated templates in our dataset were then thoroughly reviewed and corrected by human annotators who are native speakers of the languages (details in Appendix \ref{app:data_const}). 



\subsection{Multilingual Bias Evaluation (MBE)}
We also evaluate MLMs with the MBE score proposed in \cite{kaneko-etal-2022-gender} containing datasets for bias evaluation in 8 high resource languages: German (de), Japanese (ja), Arabic (ar), Spanish (es), and Mandarin (zh) belonging to Class 5; Portuguese (pt) and Russian (ru) in Class 4; and Indonesian (id) in Class 3. For evaluation, it first considers parallel corpora from English to different languages and extracts the set of sentences containing male and female words.
Next, the likelihood for each sentence is evaluated with the MLM, and the bias score is measured as the percentage of total pairs for which a male sentence gets a higher likelihood than a female sentence. Hence a value close to 50 for an MLM indicates no bias towards both groups while greater or smaller values indicate a bias towards females and males respectively. For better interpretability of metrics, we report $|50 - \mathrm{MBE}|$ in our results. 



\section{Mitigating Bias in Multilingual Models}

We next discuss how we extend bias mitigation techniques to work beyond English along with different fine-tuning and prompting strategies that we deploy in our experiments.


\subsection{Counterfactual Data Augmentation (CDA)}

CDA \cite{zhao-etal-2018-gender} is an effective method for reducing biases picked up by the language models during pre-training. It operates by augmenting an unlabeled text corpus with counterfactuals generated for each sentence based on a specific dimension like gender. As an example, the counterfactual for a sentence $s =$ {\textit{``The doctor went to \textcolor{red}{\textbf{his}} home''}} will be $\hat{s} =$\textit{``The doctor went to \textcolor{red}{\textbf{her}} home''}. The model is then fine-tuned on the augmented data, which helps balance out any spurious correlations that would have existed in the pre-training dataset.


To generate counterfactuals in English, we do word replacements on Wikipedia data using 193 gendered term pairs (eg. \{he, she\}, \{actor, actress\}, etc.) following \citet{lauscher-etal-2021-sustainable-modular}. However, generating counterfactuals for languages other than English can be challenging as acquiring term pairs need recruiting annotators which can be expensive for low-resource languages. Further, word replacement can prove unreliable for languages that mark gender case to objects (like Hindi), producing ungrammatical sentences \cite{zmigrod-etal-2019-counterfactual}.


\noindent
\\
\textbf{Generating Multilingual Counterfactuals}: We use a translation-based approach to obtain counterfactually augmented examples in different languages. We first select the sentences in the Wikipedia English corpus containing India-related keywords which were extracted using ConceptNet \cite{Speer_Chin_Havasi_2017} which include keywords related to Indian food, location, languages, religions, etc. Using these keywords we select a set of 20K sentences to avoid under-representation of Indian culture specific context. Also, generating counterfactuals for the whole corpus and fine-tuning MLMs for each of the languages will require substantial energy consumption \cite{strubell-etal-2019-energy}, so we decided to use the set of filtered 20k sentences for debiasing the MLMs. Further, we augment the 193 term pairs list to contain pairs of Indian personal names as well. We align the male and female names through a greedy search for selecting pairs with minimum edit distance. Finally, using the augmented term pairs list and the filtered data with Indian context, we generate counterfactuals using word replacements and translate the obtained data to the 6 Indian languages.

Once we have obtained CDA data in different languages, we can utilize it to debias the model. We define CDA-$\mathcal{S}$ as a fine-tuning setup where the MLM is debiased using CDA data for languages belonging to the set $\mathcal{S} \subset \mathcal{L}$, where $\mathcal{L} = \{\mathrm{en}, \mathrm{hi}, \mathrm{pa}, \mathrm{bn}, \mathrm{ta}, \mathrm{gu}, \mathrm{mr}\}$.
In particular, we explore the following classes of fine-tuning setups:

\noindent
\textbf{1. \zscda }: Fine-tune the model with English CDA data only (zero-shot debiasing).

\noindent
\textbf{2. \monocda }: Fine-tune the model with language $l$ specific CDA data (monolingual-debiasing).

\noindent
\textbf{3. \fscda }: Fine-tune the model with English and language $l$'s CDA data (few-shot debiasing).


\noindent
\textbf{4. \indcda}: Fine-tune the model with CDA data in all non-English languages (multilingual-debiasing).

\subsection{Self-Debiasing}
Self-Debiasing \cite{schick-etal-2021-self} is a post-hoc method to reduce corpus-based biases in language models. It is based on the observation that pretrained language models can recognize biases in text data fairly well and prepends the input text with prompts encouraging the model to exhibit undesired behavior. Using this, it recognizes the undesirable predictions of the model as the ones with an increase in likelihood when the prompt is provided and suppresses them in the final predictions. 
 We translate the English prompt \textit{``The following text discriminates against people because of their gender''} in different languages and use them for bias mitigation (\textbf{SD-$l$}). {We also experiment with using English prompt for other languages (\textbf{SD-$\mathrm{en}$})}.



\section{Results}

\begin{table*}[]
\small
\centering
\begin{tabular}{c c c c c c c c c c c} 

    \toprule
    MLM & Method & Languages & en & hi & pa & bn & ta & gu & mr & $\mathcal{L} \setminus \{\mathrm{en}\} $ \\
    \midrule
    \multirow{7}{*}{XLM-R} & OOB & $\{\}$ & 0.78 & 0.83 & 0.92 & 0.94 & 0.94 & 0.86 & 0.86 & 0.89 \\
    \cmidrule{2-11}
    & \multirow{1}{*}{Self-Debiasing}& $\{\mathrm{en}\}$ & 0.82 & 0.88 & 0.92 & 0.93 & 0.94 & 0.86 & 0.87 & 0.90 \\
    & & $\{l\}$ & 0.82 & 0.89 & 0.93 & 0.94 & 0.92 & 0.89 & 0.88 & 0.91 \\
    \cmidrule{2-11}
    & \multirow{4}{*}{CDA} & $\{\mathrm{en}\}$ & \textbf{0.61} & 0.83 & 0.83 & 0.89 & 0.90 & 0.82 & 0.83 & 0.85 \\
    & & $\{l\}$ &\textbf{0.61} & 0.81 & 0.84 & 0.90 & 0.92 & 0.78 & 0.83 & 0.85\\
    & & $\{l, \mathrm{en}\}$ & - & \textbf{0.74} & 0.79 & 0.88 & 0.87 & \textbf{0.70} & \textbf{0.69} & 0.78\\
    & & $\mathcal{L} \setminus {\mathrm{en}} $ & 0.73 & 0.75 & \textbf{0.61} &\textbf{ 0.87} & \textbf{0.87} & 0.78 & 0.76 & \textbf{0.77} \\
    \midrule
    \multirow{7}{*}{IndicBERT} & OOB & $\{\}$ & \textbf{0.70} & 0.79 & 0.84 & 0.93 & 0.86 & 0.82 & 0.76 & 0.83\\
    \cmidrule{2-11}
    & \multirow {1}{*}{Self-Debiasing} & $\{\mathrm{en}\}$ & 0.78 & 0.86 & 0.93 & 0.98 & 0.93 & 0.86 & 0.87 & 0.90\\
    & & $\{l\}$ & 0.78 & 0.86 & 0.89 & 0.96 & 0.91 & 0.84 & 0.87 & 0.89 \\
    \cmidrule{2-11}
    & \multirow{4}{*}{CDA} & $\{\mathrm{en}\}$ & 0.70 & 0.76 & \textbf{0.72} & 0.95 & 0.89 & 0.83 & 0.85 & 0.83 \\
    & & $\{l\}$ & 0.70 & 0.80 & 0.80 & 0.82 & 0.90 & 0.79 & 0.78 & 0.82\\
    & & $\{l, \mathrm{en}\}$ & - & 0.75 & 0.80 & 0.83 & 0.80 & 0.86 & 0.75 & 0.80 \\
    & & $\mathcal{L} \setminus {\mathrm{en}} $ & 0.72 & \textbf{0.66} &0.75 & \textbf{0.80} & \textbf{0.79} & \textbf{0.66} & \textbf{0.73} & \textbf{0.73}\\
    
    \bottomrule
    

\end{tabular}
\vspace{-2mm}
\caption{Multilingual DisCo metric results (score of 1 being fully biased and 0 being fully unbiased) of debiasing using CDA and Self-Debiasing using various fine-tuning settings on different languages. Refer to Table \ref{tab:disco_results_complete} for the full version of the results.}
\label{tab:disco_results}

\end{table*}

We evaluate the Out Of Box (OOB) biases as well the effect of applying aforementioned debiasing techniques in multilingual MLMs like XLMR-base \cite{conneau-etal-2020-unsupervised}, IndicBERT \cite{kakwani-etal-2020-indicnlpsuite}, and mBERT (cased) \cite{devlin-etal-2019-bert} using our multilingual DisCo metric. Additionally, we also evaluate language-specific monolingual models (refer Table \ref{tab:mlms_info} in appendix) and XLMR on the MBE score.

\noindent
\\
\textbf{Comparison Between Different Fine-tuning Setups for CDA}: We first compare the results of bias mitigation across all 4 classes of finetuning setups for CDA to understand the effect each had on the final bias reduction. As can be seen in Table \ref{tab:disco_results} even though zero-shot transfer from English (\zscda) results in some reduction in biases when compared to the models without any debiasing (OOB), most of the other fine-tuning setups that use language-specific counterfactuals incur better drops in the DisCo score. Specifically, few-shot debiasing (\fscda) and multilingual-debiasing (\indcda) perform consistently the best for both models with \indcda\ performing slightly better for XLMR and substantially so for Indic-BERT. This shows that even though language-specific counterfactuals were translated, using them for the debiasing of models helped in considerable bias reduction. We also observe that the monolingual debiasing (\monocda) leads to a drop similar to \zscda, and we conjecture that it might be attributed to the low amount of data we have in languages other than English for debiasing. Further, the dominant performance of \indcda\ highlights that languages from a similar culture can collectively help improve biases in such models. We also observe similar results for mBERT which are provided in Table \ref{tab:disco_results_complete} in the appendix. 


\noindent
\\
\textbf{Comparison Between CDA and Self-Debiasing}:
Counter to CDA, Self-Debiasing shows different bias mitigation trends for Indian languages. Table \ref{tab:disco_results} shows that for both multilingual MLMs, the overall bias ends up increasing when Self-Debiasing is applied, and that too by a considerable amount for IndicBERT. 
This seems to be in contrast to the past work \cite{meade-etal-2022-empirical} that shows Self-Debiasing to be the strongest debiasing technique. However, we will see next the cases where it can indeed be effective in reducing biases.

\begin{figure}
    \centering
     \begin{subfigure}{.45\textwidth}
        \includegraphics[width=.99\textwidth]{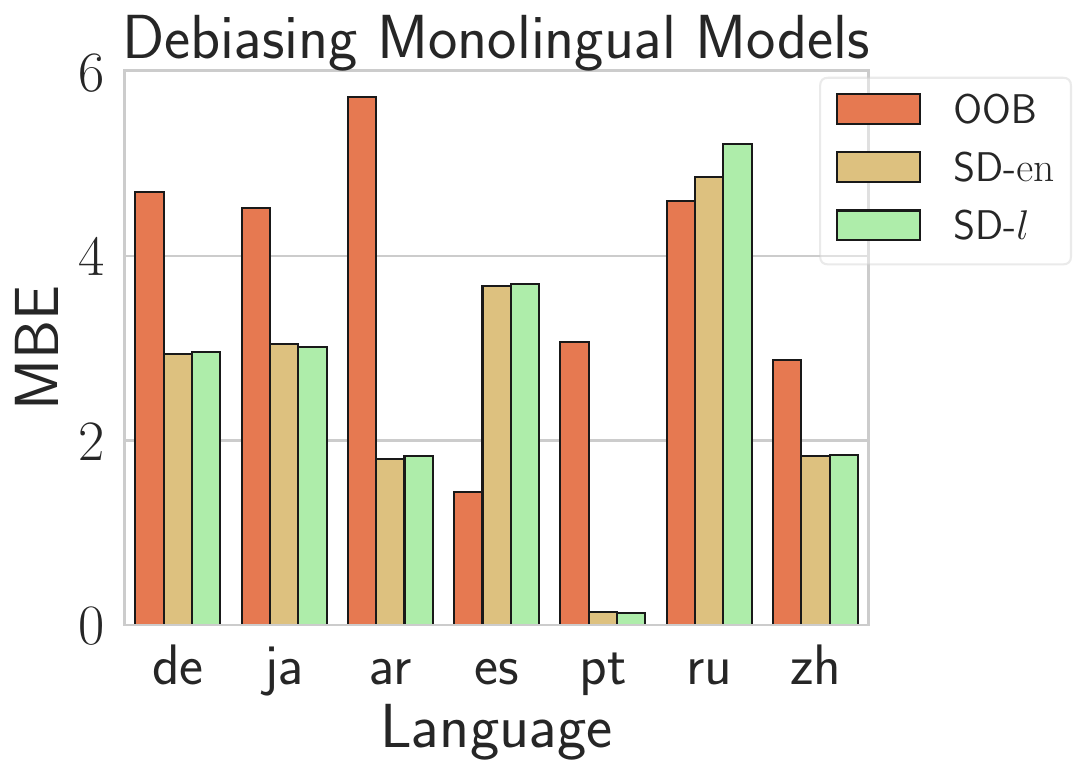}
        \caption{}
        \label{fig:monolingual_mbe}
    \end{subfigure}\\
     \begin{subfigure}{.45\textwidth}
        \includegraphics[width=.99\textwidth]{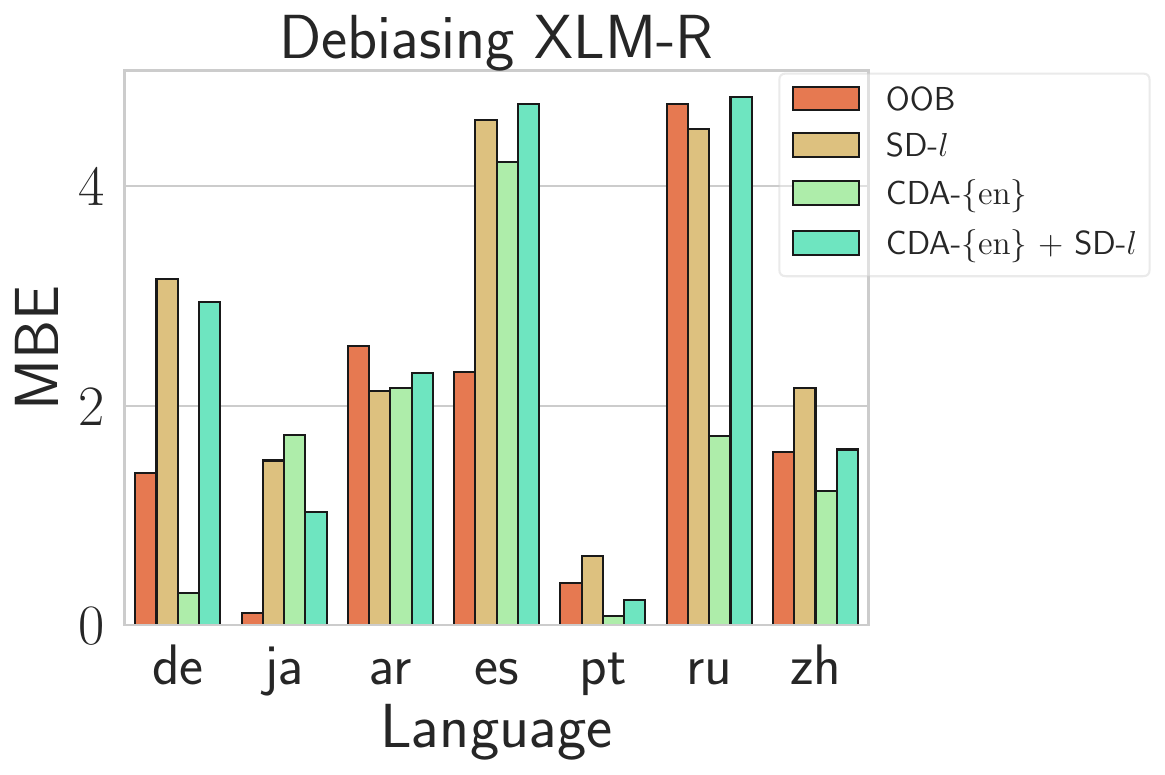}
        \caption{}
        \label{fig:xlmr_mbe}
    \end{subfigure}
    \caption{MBE scores for monolingual and multilingual models and the impact of debiasing across languages}
    \label{fig:mbe}
\end{figure}

\noindent
\\
\textbf{Evaluation on MBE Metric}:
We first investigate the effect of Self-Debiasing on monolingual models when evaluated for the MBE metric. As can be observed in Figure \ref{fig:monolingual_mbe}, for most languages (except Russian and Spanish), both variants of Self-Debiasing manage to reduce the biases substantially. However, when we compare the results on a multilingual model i.e. XLMR in Figure \ref{fig:xlmr_mbe}, we again observe the same phenomenon as for multilingual DisCo, where the biases tend to increase upon applying Self-Debiasing. Figure \ref{fig:monolingual_mbe} shows that SD-en and SD-l have similar debiasing performance for monolingual models. It is intriguing that monolingual models are able to debias so well based on English prompts. This similarity in results with non-English and English prompts could possibly be explained by contamination in the pretraining monolingual data \cite{blevins2022language}. 
We also compare the effect of \zscda on reducing the biases and we observed it does obtain more success in most languages (except Spanish and Japanese). Even though MBE and Multilingual DisCo have different experimental setups, obtaining consistent results while using the two different metrics like English-only debiasing being insufficient to reduce biases in other languages. Self-debiasing being ineffective for mitigating biases in multilingual models strenghtens the applicability of our results.
Our results indicate that Self-Debiasing might be limited for multilingual models and we leave the investigation of this phenomenon to future work.

\section{Conclusion}
In this work, we investigated gender biases in multilingual settings by proposing a bias evaluation dataset in 6 Indian languages. We further extended debiasing approaches like CDA and Self-Debiasing to work for languages beyond English and evaluated their effectiveness in removing biases across languages in MLMs. One of our key findings is that debiasing with English data might only provide a limited bias reduction in other languages and even collecting a limited amount of counterfactual data through translation can lead to substantial improvements when jointly trained with such data from similar languages. Finally, we showed that despite being effective on monolingual models, Self-Debiasing is limited in reducing biases in multilingual models with often resulting in an increase in overall bias. We hope that our work will act as a useful resource for the community to build more inclusive technologies for all cultures.

\section{Limitations}
The present study is limited to exploring biases in MLMs for the gender dimension only. For future work, important dimensionalities can be explored, especially for non-western contexts like Caste, Ethnicity, etc \cite{ahn-oh-2021-mitigating, bhatt-etal-2022-contextualizing}.
We also used Machine Translation on English counterfactuals to obtain CDA data in each language in our dataset. Translations are prone to errors and issues like \textit{Translaionese} \cite{gellerstam-1986-translationese}, especially for the lower resource languages, and therefore can lead to the unreliability of the quality of generated counterfactuals  were generated. In the future, we would like to explore learning generative \cite{wu-etal-2021-polyjuice} or editing models \cite{malmi-etal-2022-text} for automatically generating gender counterfactuals given text data in different languages. This can help us scale our counterfactual generation process to a much higher number of samples while also avoiding any losses in quality that may arise due to machine translation. Our multilingual DisCo metric is currently limited to 6 Indian languages and we hope our work will inspire further extension to cover different language families for improving the focus on multilingual biases evaluation. 
\section{Ethical Considerations}
Our work dealt with evaluating biases in MLMs and different methods for bias mitigation in multilingual settings. While most of the current work is disproportionately in favor of high-resource languages like English, it is extremely important to improve this linguistic disparity for building inclusive and responsible language technology. Through our work, we provided a dataset to evaluate gender biases in languages of varying resources as well as methods to reduce such biases.

\section*{Acknowledgements}
We would like to thank the following people who helped in evaluating and improving the Multilingual DisCo templates: Ranajoy Sadhukhan, Atharv Sonwane, Abhinav Rao, Krut Patel and Mirza Baig.

\bibliography{anthology,custom}
\bibliographystyle{acl_natbib}

\appendix
\section{Appendix}
\subsection{Dataset Construction Details}
\label{app:data_const}
\noindent
\textbf{Scraping Langauge-Specific Personal Names}: We curated a list of personal names corresponding to the cultures for each language by scraping the popular surnames associated with each culture from Wikipedia\footnote{\url{https://en.wikipedia.org/wiki/Category:Indian_surnames}}. We then obtain the open source list of Indian male\footnote{\url{https://gist.github.com/mbejda/7f86ca901fe41bc14a63}} and female\footnote{\url{https://gist.github.com/mbejda/9b93c7545c9dd93060bd}} names, and we segment the names to different languages by referring to our culture-specific surnames list. The names obtained this way our in Latin script, so we transliterate them to the corresponding languages using the Bing Translator API.

\noindent
\textbf{Annotator Details}: For verifying the templates obtained using machine translation we asked human annotators to correct them. Our annotators were colleagues working at our research lab and all of them were of South Asian (Indian) descent, native to different parts of India, and each having one of the six Indian languages that we consider as their L1. They all identify as males and are in their mid-20s. 
\noindent
The annotators were provided original English templates along with the translated ones in their native language and were asked to verify that they were grammatically correct and conveyed the exact same meaning as the original base template. Further, they were asked to make corrections to ensure that a template pair was as close to each other as possible except for modifications in the gendered terms, like verbs in the case of Hindi (Figure \ref{fig:mbe}).

\noindent
\textbf{Dataset Statistics:}
Our dataset consists of 14 templates in each language and for each language the number of name pairs are given in Table \ref{tab:name_pair_stats}. 

\begin{table}
\centering
\begin{tabular}{c c} 
 \toprule
 \textbf{Language} & \textbf{Number of Name Pairs} \\ 
 \midrule
 Hindi & 164  \\ 
 \midrule
 Punjabi & 50  \\
 \midrule
 Bengali & 33  \\
 \midrule
 Gujarati & 51 \\
 \midrule
 Tamil & 19  \\
 \midrule
 Marathi & 49  \\
 \bottomrule
\end{tabular}
\caption{Total number of gendered name pairs for each language used in Multilingual DisCo}
\label{tab:name_pair_stats}
\end{table}

\subsection{Experimental Setup}
\label{app:expt_setup}
We performed all our experiments on a single A100 GPU. For the fine-tuning setup \zscda, we trained for 50K steps using a batch size of 32, a learning rate of 2e-5, and a weight decay of 0.01. We follow the same hyperparameters for other fine-tuning setups as well, but instead of fine-tuning for 50K steps, we train for 1 epoch following \cite{lauscher-etal-2020-zero} as the amount of data is limited in other languages. For Self-Debiasing, we used the default hyperparameters i.e. the decay constant $\lambda = 50$ and $\epsilon = 0.01$. For all of our experiments, we used the pre-trained models provided with HuggingFace's transformers library \cite{wolf-etal-2020-transformers}. The details of all the pre-trained models that we use in the paper are provided in Table \ref{tab:mlms_info}
\begin{table*}[]
    \centering
    \resizebox{\linewidth}{!}{%
    \begin{tabular}{lccc}
         \toprule
         \textbf{Model Name} & \textbf{Variant} & \textbf{Supported Languages} & \textbf{Number of Parameters} \\
         \midrule
         \multicolumn{4}{l}{\textit{\textbf{Multilingual Masked Language Models}}}\\
         \midrule
         XLM-R & xlm-roberta-base & 100 languages from \cite{conneau-etal-2020-unsupervised} & 270M \\
         \midrule
         IndicBERT & {indic-bert} & 12 Indian Languages & 12M \\
         \midrule
         mBERT & bert-base-multilingual-cased & Top 104 Wikipedia Languages \footnote{\url{https://github.com/google-research/bert/blob/master/multilingual.md}} & 110M\\
         \midrule
         \multicolumn{4}{l}{\textit{\textbf{Monolingual Masked Language Models}}}\\
         \midrule
        GBERT \cite{chan-etal-2020-germans} & gbert-base & de & 110M \\
         \midrule
         BERT Japanese\footnote{\url{https://github.com/cl-tohoku/bert-japanese/tree/v1.0}} & bert-base-japanese-whole-word-masking & ja& 110M \\
         \midrule
         AraBERT \cite{antoun-etal-2020-arabert} & bert-base-arabertv02 & ar & 110M \\
         \midrule
         Spanish Pre-trained BERT \cite{CaneteCFP2020} & {bert-base-spanish-wwm-uncased} & es & 110M \\
         \midrule
         BERTimbau\cite{souza2020bertimbau}  & {bert-base-portuguese-cased} & pt &  110M \\
         \midrule
         RoBERTa-base for Russian \footnote{\url{https://huggingface.co/blinoff/roberta-base-russian-v0}} & {roberta-base-russian-v0} & ru & 110M \\
         \midrule
         Chinese BERT \cite{cui-etal-2020-revisiting} & {chinese-bert-wwm-ext} & zh & 100M\\
         \bottomrule
    \end{tabular}}
    \caption{Description of MLMs that we use in our experiments}
    \label{tab:mlms_info}
\end{table*}

\begin{table*}[]
\small
\centering
\begin{tabular}{c c c c c c c c c c c}
    \toprule
    MMLM & Debiasing Method & Languages Used & {en} &  hi & pa & bn & ta & gu & mr & $\mathcal{L} \setminus \{\mathrm{en}\} $ \\
    \midrule
    \multirow{8}{*}{XLM-R} & OOB & $\{\}$ & 0.78 & 0.83 & 0.92 & 0.94 & 0.94 & 0.86 & 0.86 & 0.89 \\
    \cmidrule{2-11}
    & \multirow{2}{*}{Self-Debiasing} & $\{\mathrm{en}\}$ & 0.82 & 0.88 & 0.92 & 0.93 & 0.94 & 0.86 & 0.87 & 0.90 \\
    & & $\{l\}$ & 0.82 & 0.89 & 0.93 & 0.94 & 0.92 & 0.89 & 0.88 & 0.91 \\
    \cmidrule{2-11}
    & \multirow{4}{*}{CDA} & $\{\mathrm{en}\}$ & \textbf{0.61} & 0.83 & 0.83 & 0.89 & 0.90 & 0.82 & 0.83 & 0.85 \\
    & & $\{l\}$ &\textbf{0.61} & 0.81 & 0.84 & 0.90 & 0.92 & 0.78 & 0.83 & 0.85\\
    & & $\{\mathrm{en}, l\}$ & - & \textbf{0.74} & 0.79 & 0.88 & 0.87 & \textbf{0.70} & \textbf{0.69} & 0.78\\
    & & $\mathcal{L} \setminus {\mathrm{en}} $ & 0.73 & 0.75 & \textbf{0.61} &\textbf{ 0.87} & 0.87 & 0.78 & 0.76 & \textbf{0.77} \\
    & & $\mathcal{L}$ & 0.72 & 0.78 & 0.74 & 0.89 & \textbf{0.85} & 0.75 & 0.79 & 0.80 \\
    \midrule
    \multirow{8}{*}{mBERT} & OOB & $\{\}$ & 0.88 & 0.87 & 0.72 & 0.93 & \textbf{0.79} & 0.84 & \textbf{0.71} & 0.81 \\
    \cmidrule{2-11}
    & \multirow{2}{*}{Self-Debiasing} & $\{\mathrm{en}\}$ & 0.88 & 0.90 & 0.87 & 0.98 & 0.94 & 0.91 & 0.89 & 0.91\\
    & & $\{l\}$ & 0.88 & 0.86 & 0.81 & 0.98 & 0.92 & 0.91 & 0.82 & 0.88\\
    \cmidrule{2-11}
    & \multirow{4}{*}{CDA} & $\{\mathrm{en}\}$ & \textbf{0.68} & 0.90 & 0.73 & 0.94 & 0.85 & 0.79 & 0.75 & 0.83 \\
    & & $\{l\}$ &\textbf{0.68} & \textbf{0.76} & 0.72 & 0.89 & 0.86 & 0.77 & 0.79 & {0.80}\\
    & & $\{\mathrm{en}, l\}$ & - & 0.84 & \textbf{0.67} & 0.86 & 0.80 & \textbf{0.73} & 0.76 & \textbf{0.78}\\
    & & $\mathcal{L} \setminus {\mathrm{en}} $ & 0.88 & 0.82 & 0.73 & \textbf{0.80} & \textbf{0.79} & 0.79 & 0.88 & {0.80} \\
    & & $\mathcal{L}$ & 0.88 & 0.83 & 0.79 & 0.81 & 0.82 & {0.75} & 0.92 & 0.82 \\
    \midrule
    \multirow{5}{*}{IndicBERT} & OOB & $\{\}$ & 0.70 & 0.79 & 0.84 & 0.93 & 0.86 & 0.82 & 0.76 & 0.83\\
    \cmidrule{2-11}
    & \multirow{2}{*}{Self-Debiasing} & $\{\mathrm{en}\}$ & 0.78 & 0.86 & 0.93 & 0.98 & 0.93 & 0.86 & 0.87 & 0.90\\
    & &  $\{l\}$ & 0.78 & 0.86 & 0.89 & 0.96 & 0.91 & 0.84 & 0.87 & 0.89 \\
    \cmidrule{2-11}
    & \multirow{4}{*}{CDA} & $\{\mathrm{en}\}$ & 0.70 & 0.76 & \textbf{0.72} & 0.95 & 0.89 & 0.83 & 0.85 & 0.83 \\
    & & $\{l\}$ & 0.70 & 0.80 & 0.80 & 0.82 & 0.90 & 0.79 & 0.78 & 0.82\\
    & & $\{\mathrm{en}, l\}$ & - & 0.75 & 0.80 & 0.83 & 0.80 & 0.86 & 0.75 & 0.80 \\
    & & $\mathcal{L} \setminus {\mathrm{en}} $ & 0.72 & \textbf{0.66} & 0.75 & \textbf{0.80} & \textbf{0.79} & \textbf{0.66} & \textbf{0.73} & \textbf{0.73}\\
    & & $\mathcal{L}$ & \textbf{0.62} & 0.73 & 0.82 & 0.85 & 0.85 & 0.79 & 0.76 & 0.80 \\
    \bottomrule
    

\end{tabular}
\caption{Complete version of results of debiasing using CDA and Self-Debiasing using various fine-tuning settings on different languages and MMLMs.}
\label{tab:disco_results_complete}
\end{table*}

\end{document}